\documentclass[sigconf]{acmart}

\usepackage[vlined]{algorithm2e}
\usepackage{amsmath}
\usepackage{amsfonts}       %
\usepackage[utf8]{inputenc} %
\usepackage[T1]{fontenc}    %
\usepackage{hyperref}       %
\usepackage{url}            %
\usepackage{nicefrac}       %
\usepackage{microtype}      %
\usepackage{amsthm}
\usepackage{subcaption}
\usepackage{rotating}
\usepackage{graphicx}

\usepackage{booktabs}       %
\usepackage{xcolor}

\usepackage[titletoc,toc,title]{appendix}

\usepackage[inline]{enumitem}
\usepackage{mathtools}

\usepackage{Definitions}

\newlist{inlinelist}{enumerate*}{1}
\setlist*[inlinelist,1]{%
  label=(\arabic*),
}

\newcommand{\bX}{\mathbf{X}}

\newcommand{\E}{\mathbb{E}}
\newcommand{\ignore}[1]{}

\setcopyright{rightsretained}

\acmArticle{4}
\acmPrice{15.00}

\renewcommand\footnotetextcopyrightpermission[1]{} %
\acmBooktitle{Proceedings of KDD'18, Aug 19-23, London, UK}

\begin{document}

\setlength{\abovedisplayskip}{3pt}
\setlength{\belowdisplayskip}{3pt}

\title{Efficient Inner Product Approximation in Hybrid Spaces}

\author{\textbf{Xiang Wu} \,\, \textbf{Ruiqi Guo} \,\, \textbf{David Simcha} \,\, \textbf{Dave Dopson} \,\, \textbf{Sanjiv Kumar} \\
Google Research, New York \\
\{\texttt{wuxiang}, \texttt{guorq}, \texttt{dsimcha}, \texttt{ddopson}, \texttt{sanjivk}\} \texttt{@google.com}
}

\begin{abstract}
Many emerging use cases of data mining and machine learning operate on large datasets with data from heterogeneous sources, specifically with both sparse and dense components.
For example, dense deep neural network embedding vectors are often used in conjunction with sparse textual features to provide high dimensional hybrid representation of documents. 
Efficient search in such hybrid spaces is very challenging as the techniques that perform well for sparse vectors have little overlap with those that work well for dense vectors. Popular techniques like Locality Sensitive Hashing (LSH) and its data-dependent variants also do not give good accuracy in high dimensional hybrid spaces. Even though hybrid scenarios are becoming more prevalent, currently there exist no efficient techniques in literature that are both fast and accurate. In this paper, we propose a technique that approximates the inner product computation in hybrid vectors, leading to substantial speedup in search while maintaining high accuracy. 
We also propose efficient data structures that exploit modern computer architectures, resulting in orders of magnitude faster search than the existing baselines.
The performance of the proposed method is demonstrated on several datasets including a very large scale industrial dataset containing one billion vectors in a billion dimensional space, achieving over 10x speedup and higher accuracy against competitive baselines.
\end{abstract}


\maketitle

\section{Introduction}
\label{sec:intro}

An astronomical amount of data has been produced by human race in recent years compared with the past history~\cite{LynchBigData}.
It is valuable to understand and process this data using data mining and machine learning.
We cannot fail to notice that the data is generated from diverse sources in practice. For example, it may come from web services, mobile devices, location services or digital imagery. The feature vectors representing these records have varying properties e.g., dimensionality (high or low), sparsity (sparse or dense) and modality (categorical or continuous). 

We are interested in developing effective methods to process, index and search large scale datasets of heterogeneous nature. Specifically, we focus on datasets consisting of feature vectors that combine sparse and dense features. Such representations are increasingly popular in several applications. For instance, documents have been traditionally represented as a vector of normalized n-grams (e.g., unigrams and bigrams). It is easy to have millions or billions of n-grams, leading to high dimensional representations. For each document only a few such n-grams are present, yielding a very sparse vector representation. On the contrary, with widespread use of deep learning techniques, it is common to represent text as embeddings e.g., word2vec~\cite{MikolovWord2Vec} or LSTM hidden states~\cite{PalangiLSTMEmbedding}. Such embeddings are typically low dimensional (in hundreds) but dense, i.e., all features are nonzero. Both n-gram (sparse) and embedding (dense) representations have pros and cons. Embeddings  capture correlations among n-gram features, which is not possible with direct n-gram representations. On the contrary, direct n-grams can "memorize" specific rare features that are helpful in document representations. To get the best of both worlds, concatenations of sparse and dense features are increasingly used in learning tasks. For instance, in a recent ``Wide-and-Deep'' framework~\cite{ChengWideAndDeep}, data has both a dense component (from the ``deep'' neural network) and a sparse component (from the ``wide'' sparse logistic regression). 
Fast computation of inner products (or equivalently cosine similarity\footnote{Cosine similarity between two vectors is equivalent to their inner product after normalizing the vectors to unit L2 norms.}) in sparse-dense hybrid spaces is useful in several applications:
\begin{itemize}
\item Finding similar items in a hybrid dataset such as DBLP~\cite{LeyDBLP};
\item Collaborative filtering with a hybrid of sparse and dense features~\cite{ChengWideAndDeep, XiaoAttentionFM};
\item Extreme classification where the classifier makes use of both dense and sparse features while the number of classes is large (commonly seen in recommendation tasks~\cite{ShanDeepCrossing,WangDCN}).
\end{itemize}

\subsection{Insufficiency of the existing methods}
\label{sec:intro_analysis}

Historically, disjoint approaches to efficient inner product search have been applied to sparse and dense datasets. The most popular way to compute inner product for high dimensional sparse vectors is based on variants of inverted indices~\cite{ManningIRBook} which exploit the sparsity of non-zero (``active'') dimensions in the data. 
However, the performance of inverted indices degrades quickly when even a few dimensions are dense. On the contrary, search in dense data is usually based on either tree-based or hashing-based techniques. For example, KD-Tree~\cite{Kdtree}, VP-Tree~\cite{yianilos1993data} and Spill-Tree~\cite{SpillTree}, have been widely used for lower data dimensionalities (typically less than one hundred). 
For higher dimensional data, hashing methods~\cite{SanjivSurvey,ALSH,FALCONN} or quantization methods~\cite{PQ,QUIPS,AQ,LOPQ,CQ,CKmeans} are more popular. 
But none of these techniques scales to billions of dimensions without significantly degrading search recall.

Another approach is to search the sparse and dense spaces separately, using appropriate data structures, and then combining the results. However, such heuristics can easily fail when the most query-similar items in the combined space are only middling in the dense and sparse spaces individually.
We also note that projection based methods such as LSH~\cite{LSH} and RPTree~\cite{RPTree} can be applied to hybrid data because they first project the data onto a low-dimensional space, but such techniques often result in sizable recall loss.

Currently, no good solutions exist for efficiently searching with high-recall in the hybrid spaces that are the focus of this paper.
We highlight the insufficiency of the existing methods by extensive experiments (Section~\ref{sec:experiments}). For example, for the large-scale QuerySim dataset consisting of a billion datapoints with 200 dense and 1 billion sparse dimensions (with about 200 nonzeros per vector), treating all the dimensions as dense is infeasbile. Moreover, treating all the dimensions as sparse leads to abysmally slow search performance
(a few minutes per query).
This is because the dense dimensions of the dataset are active in all vectors, leading to full inverted lists for these dimensions. Similarly, projection related methods like LSH give very poor performance even with a large number of bits.

\subsection{Contributions}
\label{sec:intro_contrib}

This paper makes the following main contributions: 
\begin{itemize}
\item introduces the challenging problem of efficient search in sparse-dense hybrid vector spaces and addresses it by developing a fast approximation of inner products in such spaces,
\item identifies the bottleneck of fast inner product approximation in sparse components, and designs a novel data structure called cache-sorted inverted index to overcome it. Such a data structure makes inner product approximation streamlined with SIMD operations on modern CPU architectures and avoids expensive cache-misses (Section.~\ref{sec:cache_sort}),
\item approximates inner product in dense components by learning data-dependent quantization in subspaces, indexed with cache-friendly Look Up Tables (LUT16) and coupled with fast SIMD computations (Section.~\ref{sec:indexing_dense}),
\item proposes residual reordering which reduces the impact of approximation error in inner products, improving the accuracy substantially (Section.~\ref{sec:residual}),
\item demonstrates the effectiveness of the proposed method on two public and one very large industrial datasets,  significantly outperforming competitive baselines (Section.~\ref{sec:experiments}).

\end{itemize}

\section{Our Approach}
\label{sec:method}

\subsection{Overview}
We start with a high-level overview of our approach. As mentioned before, we address the challenge of efficient search in hybrid spaces via fast approximation of inner products. Due to the decomposibility of the inner product measure, one can approximate the inner product in sparse and dense spaces separately. Formally, let us consider a hybrid vector $x \in R^d$, with sparse subvector $x^S \in R^{d^S}$ and dense subvector $x^D \in R^{d^D}$, where $d = d^S + d^D$ and $x = x^S \oplus x^D$. 
Given a query $q$, the inner product of $q$ and $x$ can be written as,
\begin{equation}
q \cdot x = q^S \cdot x^S + q^D \cdot x^D,
\end{equation}
where we refer to $q^S \cdot x^S$ and $q^D \cdot x^D$ as sparse and dense inner product, respectively. Similarity search in this paper refers to finding item(s) from a hybrid dataset $\bX$ with highest inner product(s) to query $q$, i.e.,
\begin{equation*}
x^* = \argmax_{x \in \bX} q \cdot x
\end{equation*}
In this work, we approximate sparse and dense inner products independently. Specifically, we propose a novel cache-sorted inverted index for sparse inner-product, and a product code based quantization method with cache-efficient LookUp Table (LUT16) for dense inner product. The errors induced due to approximate inner-product computations are alleviated by using a second pass reordering with  residual indices (Section~\ref{sec:residual}), leading to substantial increase in recall. We start with a description of inverted index and product codes based inner product computation below. The proposed efficient data structures are described in Section~\ref{sec:ds}.  

\subsection{Inverted Index for Sparse Inner Product}
\label{sec:method_sparse}
Given a set of $N$ sparse vectors $\bX^S$ consisting of just the sparse dimensions $\mathcal{S}$ of the hybrid set $\bX$, an inverted index  constructs $d^S$ inverted lists, each corresponding to a dimension from the sparse dimension set $\mathcal{S}$. The $j^{th}$ list consists of indices of all the datapoints whose $j^{th}$ dimension is nonzero, and whose corresponding values for this dimension, i.e., $I_j = \{(i, \bX^{Si}_{j}) : \bX^{Si}_{j}\neq 0, j \in \mathcal{S}\}$, 
where $\bX^{Si}_{j}$ implies $j^{th}$ element of the $i^{th}$ vector in $\bX^S$. The complete inverted index is defined by $\mathbf{I}=\{I_j : j \in \mathcal{S}\}$. The inner product of sparse component of query, $q^S$ with the $i^{th}$ vector of dataset $\bX^S$ can be computed by summing over the nonzero dimensions of $q^S$, i.e., $j\in nz(q)$: 
\begin{equation*}
\label{eqn:inverted_index}
q^{S} \cdot \bX^{Si} = \sum_{j} q^S_j w_{ij};~\text{for}~~~(i, w_{ij}) \in I_j
\end{equation*}

\begin{figure}
\centering
\includegraphics[width=1.05 \columnwidth]{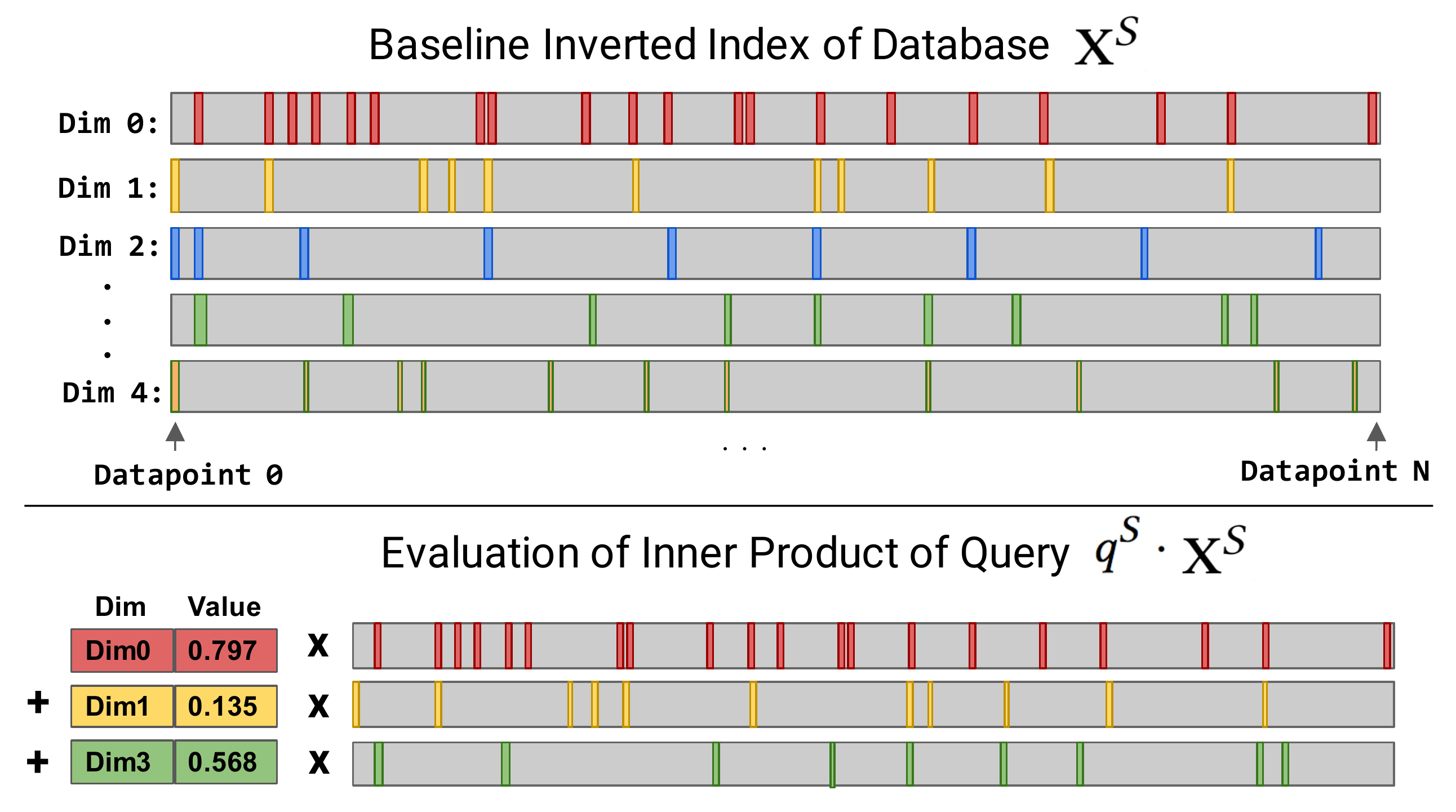}
\caption{Basics of inverted index for indexing sparse component of the dataset. Essentially, an inverted index keys the dataset by nonzero dimensions. Inner products of a query $q$ with all $x\in \{\bX^i\}$ are computed by accumulating partial inner products of $q^S_j$ and $I_j$ as described in Section~\ref{eqn:inverted_index}. Each small tile indicates nonzero values from datapoints, and the color codes indicate values on the same dimension.}
\vspace{-0.1in}
\label{fig:inverted_index}
\end{figure}

While iterating over each nonzero dimension $j$ of $q$, one can accumulate the inner products for all points in set $\bX^S$ that are active in the $j^{th}$ inverted list as illustrated in Figure~\ref{fig:inverted_index}. This accumulation based approach was first suggested in information retrieval literature~\cite{wong1993implementations,moffat1996self,zobel2006inverted}, and also employed by~\cite{BayardoAllPair}. 

Although the idea of inverted index is fairly simple, it is non-trivial to implement a practically efficient solution. Several implementation heuristics have been proposed in the  literature~\cite{wong1993implementations,moffat1996self,zobel2006inverted} but they are mostly based on old computer architectures such as slowness of floating point operations which is no longer an issue with modern CPUs. Another line of heuristics was suggested with regard to making disk-based access more efficient, which is again not relevant to modern distributed systems that tend to use in-memory inverted indices. In this paper, we claim that the major bottleneck in practical inverted index implementation is memory access and propose a new technique called \textit{cache-sorting} which can give several-fold gains in sparse inner product computation. More details are given in Section \ref{sec:cache_sorting}.

\subsection{Product Codes for Dense Inner Product}
\label{sec:pq_intro}

For indexing high-dimensional dense data, product codes based on quantization in subspaces~\cite{VQBook2} have become very popular. Product codes were used for fast Euclidean distance based search by \cite{PQ}, which along with its variants have regularly claimed top spots on public benchmarks such as GIST1M, SIFT1B~\cite{PQ} and DEEP1B~\cite{Deep10m}. Product codes were later extended to the inner-product similarity measure by \cite{QUIPS}, which showed dramatic improvement in performance over LSH based techniques. Recently product codes have also been used for  general matrix compression in \cite{BlalockBOLT}.

We approximate the inner products for the dense component of  hybrid vectors via product quantization similar to \cite{QUIPS}. The dense dimensions are split into $K$ blocks of contiguous subvectors and vector quantization is applied on each subspace separately.
Here, \textit{Vector Quantization} (VQ) approximates a $p$ dimensional vector $x$ by finding its closest quantizer in a codebook $\mathbf{U}$:

\begin{equation}
\phi_{VQ}(x; \mathbf{U}) = \argmin_{u \in \mathbf{U}} \| x-u \|_2 \nonumber
\end{equation}
where $\mathbf{U} \in \RR^{p \times l}$ is the quantization codebook with $l$ codewords.

To index a hybrid vector's dense component $x^D$, we first decompose it into $K$ subvectors, leading to its product quantization:
\begin{equation}
\phi_{PQ}(x^D; \{\mathbf{U}^{(k)}\}_{[K]}) = [\phi_{VQ}(x^{D(1)}; \mathbf{U}^{(1)}); \cdots; \phi_{VQ}(x^{D(K)}; \mathbf{U}^{(K)})]
\end{equation}
where $x^{D(k)}$ denotes the $k^{th}$ subvector of $x^D$, $\{\mathbf{U}^{(k)}\}_{[K]}$ is a collection of $K$ codebooks, and $\mathbf{U}^{(k)} \in \RR^{\textrm{dim}(x^{D(k)}) \times l}$ is the $k^{th}$ PQ codebook with $l$ sub-quantizers. The codebooks are learned using $k$-Means in each subspace independently \cite{QUIPS}. 

\begin{figure}
\centering
\includegraphics[width=1.05 \columnwidth]{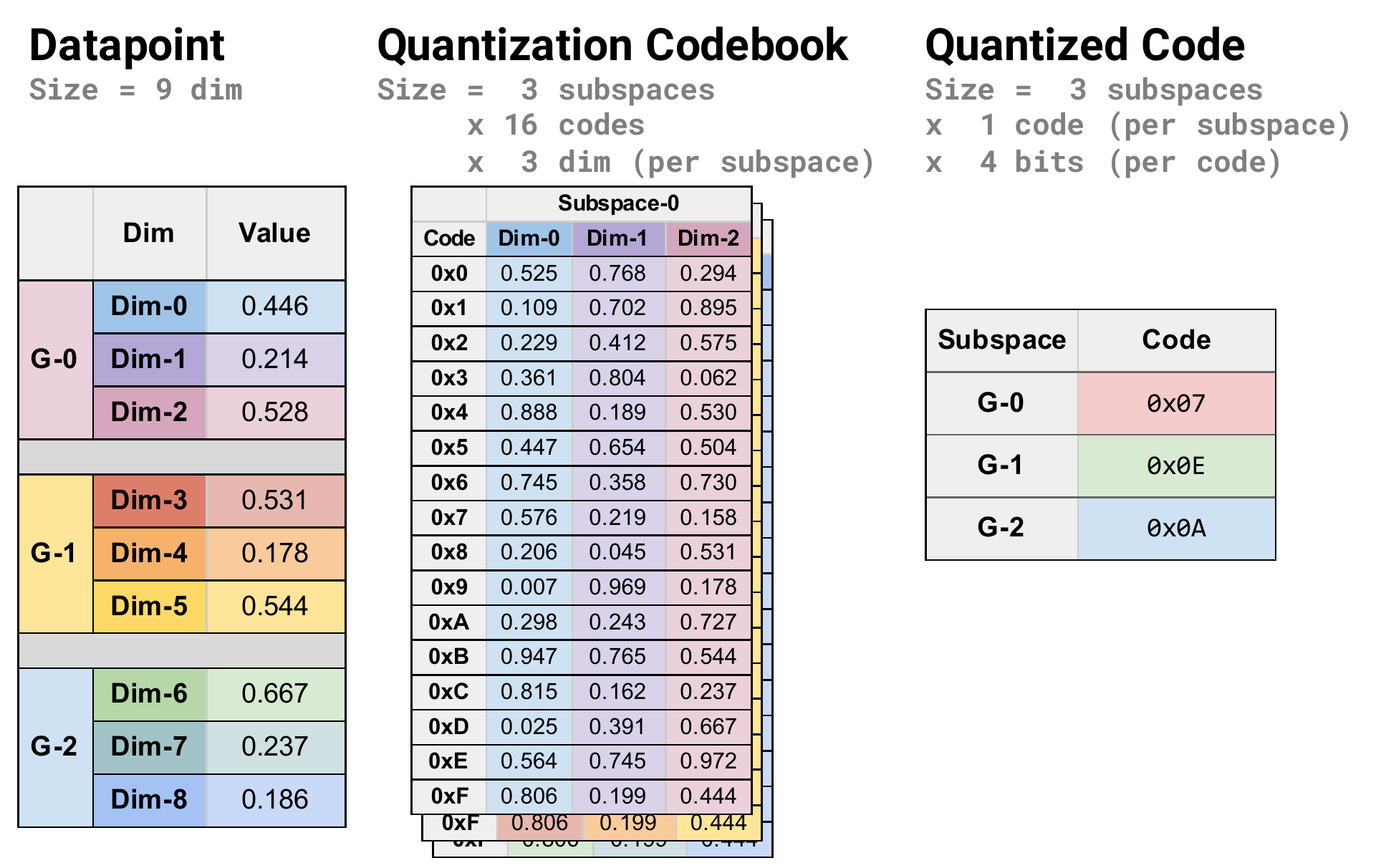}
\caption{An illustration of indexing the dense component of the dataset using a set of learned quantization codebooks $\{\mathbf{U}^{(k)}\}_{[K]}$, one for each subspace. The resulting representation contains the $K$ codes, one for each subspace. In above example, $K$ = 3, $l$ = 16 (4 bits/code). The color codes indicate matching dimensionality.}
\vspace{-0.15in}
\label{fig:pq}
\end{figure}

The indexing (quantization) procedure is illustrated graphically in Figure~\ref{fig:pq} and the dense inner product is then approximated as inner product of $q^D$ and the product quantization of $x^D$:
\begin{equation}
\label{eqn:pq}
q^{D} \cdot x^D \approx q^{D} \cdot \phi_{PQ}(x^D; \{\mathbf{U}^{(k)}\}_{[K]})
\end{equation}

\section{Efficient Data Structures}
\label{sec:ds}

In addition to asymptotic complexity, the success of an algorithm often depends upon efficient implementation on modern computing architectures, where memory bandwidth, cache access patterns, and utilization of instruction level parallelism (ILP) can have order-of-magnitude impact on performance. 
Well implemented algorithms can sometimes even outperform algorithms with lower asymptotic complexity (the well known example being matrix multiplication).

We found these factors to be critical to inner product search.
Although we have described our formulation for indexing hybrid datasets, in practice an efficient implementation is non-obvious. 
In this section, we focus on the {\em computational} aspects of implementing hybrid inner product search.

\subsection{Memory I/O and Inverted Index}
\label{sec:cache_sort}

Modern x86 processors arrange memory into a sequence of 64-byte ``cache-lines''.  When a program reads from a memory location that is not already cached, the CPU loads the 64-byte cache-line containing that address. The CPU never reads or writes fewer than 64 bytes at a time, even if the program only utilizes a small portion of that cache-line.

In practice, query performance for sparse inverted indices is significantly more constrained by the memory bandwidth required to access the accumulator than by the rate at which the CPU performs arithmetic operations. Simply counting the expected number of cache-lines touched per query provides an accurate estimation of query time.

Each accumulator cache-line can hold a fixed number of values $B$; on x86, $B$ is $16$  for 32-bit accumulators, and $32$ for 16-bit ones.
Within a given dataset, each aligned block of $B$ consecutive datapoints shares an accumulator cache-line.
For a particular dimension, if any of these $B$ datapoints is nonzero, all queries active in that dimension will have to access the corresponding cache-line.
The cost of processing a second datapoint within the same cache-line is negligible compared to the cost of accessing another cache-line.

The expected query cost can be estimated as follows.  For a sparse dataset, $X^S$, containing $N$ datapoints, where $X^{Si}_j$ is the value of the $i^{th}$ datapoint on the $j^{th}$ dimension, and where $Q_j$ is the probability that the $j^{th}$ dimension is active in a randomly sampled query vector:

$$
  \textrm{Cost}(X^S) \approx
  \sum_{j} 
  \sum_{b=0}^{N/B - 1} 
  \begin{cases}
    Q_j,& \text{if } 0 <\sum_{i=B\times b}^{B\times(b+1) - 1} \mathbb{I}(\bX^{Si}_j \neq 0) \\
    0,& \text{otherwise}
  \end{cases}
$$

Given a sparse dataset, $\bX^S$, our goal is to find a permutation $\pi$ for the ordering of datapoints which minimizes Cost($X^S$), a process that we call {\em cache sorting}.

\begin{figure}
\centering
\includegraphics[width=1.05 \columnwidth]{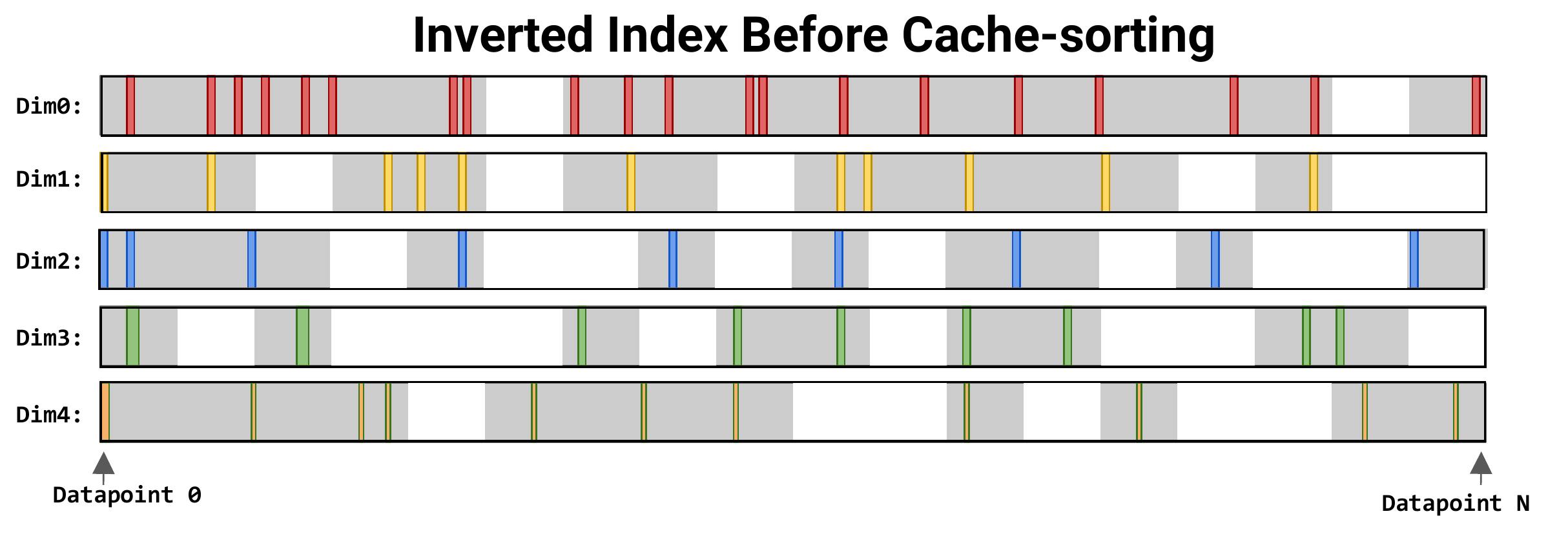}
\includegraphics[width=1.05 \columnwidth]{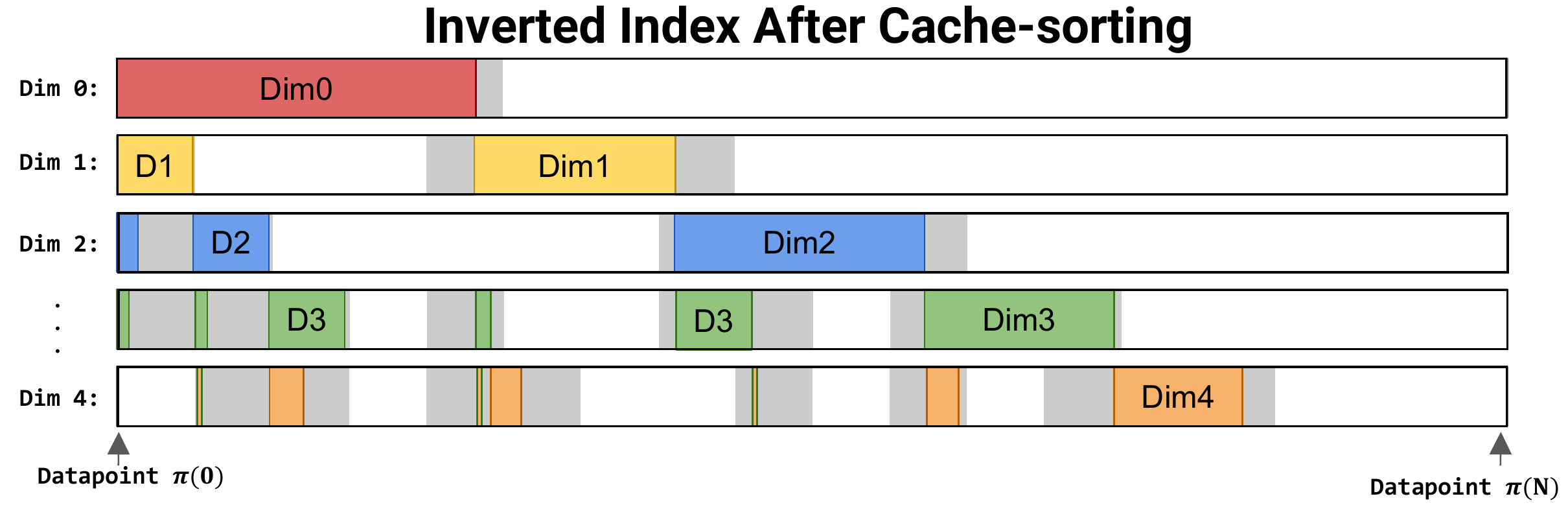}
\caption{Motivation of cache sorted inverted index. The white color indicates that a cache-line can be skipped altogether. Note that after sorting, many more cache-lines can be skipped and those accessed are also more sequential.}
\label{fig:cache_sorting}
\end{figure}

\subsection{Cache Sorting}
\label{sec:cache_sorting}

\IncMargin{-0.5em}
\begin{algorithm}
\SetKwFunction{FFCacheSort}{CacheSort}
\SetKwFunction{FFArgPartition}{Argpartition}
\SetKwFunction{FFArgSort}{Argsort}
\SetKwFunction{FFPivot}{pivot}
\SetKwFunction{FFStart}{start}
\SetKwFunction{FFEnd}{end}
\SetKwFunction{FFPartition}{PartitionByDim}
\SetKwInOut{Input}{input}
\SetKwInOut{Output}{output}
\DontPrintSemicolon

\FFCacheSort{$\bX^S$} \Begin {
  \Input{A sparse dataset $\bX^S \in \mathbb{R}^{N \times d^S}$}
  \Output{A permutation $\pi: [N] \mapsto [N]$}
  \BlankLine
  \For{$j \leftarrow 1$ \KwTo $d^S$} {
    $nnz_j \leftarrow \sum_{i=1}^N \mathbb{I}(\bX^{Si}_j \neq 0) $
  }
  \tcp{Sort dimensions by number of nonzeros.}
  $\eta \leftarrow  \FFArgSort(nnz)$   \;
  \tcp{Initialize $\pi$ to be identity.}
  $\pi \leftarrow 1, 2, \cdots, N$ \;
  \Return \FFPartition($\bX^S, 1, N, 1, \pi, \eta$)
}

\FFPartition{$\bX^S, \FFStart, \FFEnd, j, \pi, \eta$}
\Begin {
  \Input{Dataset $\bX^S$, index range to be sorted [\FFStart, \FFEnd], \\
            Dimension to perform partition $j$, \\
            Sorted ordering of the dimensions $\eta$}
  \Output{Permutation in range [\FFStart, \FFEnd]}
  \BlankLine
  \If{\FFEnd - \FFStart $\leq 1$}{
    \Return [\FFStart $\cdots$ \FFEnd]
  }
  \BlankLine
  \tcp{Partition the range by $\eta(j)$.}
  \For{$i \leftarrow$ \FFStart \KwTo \FFEnd} {
  $v_i \leftarrow \mathbb{I}(\bX^{S\pi(i)}_{\eta(j)} \neq 0)$
  }
  $\pi'$, \FFPivot $\leftarrow$ \FFArgPartition($\{v_i\}_{i \in [\FFStart, \cdots , \FFEnd]}$) \;
  \BlankLine 
  \tcp{Recursively partition the subpartitions \\by the next most active dimension.}
  $\pi_1 \leftarrow \FFPartition(\bX^S, \FFStart, \FFPivot-1, j+1, \pi \circ \pi', \eta)$ \\
  $\pi_0 \leftarrow \FFPartition(\bX^S, \FFPivot, \FFEnd, j+1, \pi \circ \pi', \eta)$ \;
  \BlankLine
  \Return $\pi' \circ [\pi_1; \pi_0]$
}

\caption{Cache sorting algorithm. This computes a permutation $\pi$ of the datapoint index, which reduces the number of main memory accesses and optimizes the access pattern. Here \texttt{Argsort} performs an indirect sort and return the indices that index the data in a sorted order. \texttt{Argpartition} performs an indirect partitioning and returns the indices and pivot.} 
\label{algo:cache_sort}
\vspace{-0.02in}
\end{algorithm}
\DecMargin{1em}

The general form of this optimization problem is known to be hard to solve.
We thus propose a greedy algorithm that produces highly efficient accumulator memory layout, even when compared with state-of-the-art approximation algorithms using LP relaxation.
Moreover, the greedy approach only takes few seconds even with millions of datapoints, while LP relaxation approaches are much more expensive.

The goal in cache sorting is to reorder the datapoint indices of shared active dimensions into long consecutive blocks as much as possible. This is done by greedily selecting the most active dimension (i.e., the dimension that is nonzero in most data vectors), and partitioning the dataset into two subsets, where the indices of the datapoints that have nonzero values for that dimension are contiguous.
This is performed recursively in the order of the dimensions, from the most active to the least active dimension. The details of the algorithm are given in Algorithm~\ref{algo:cache_sort}.

Define $\eta$ as the permutation on dimensions that sorts them from the most active to the least, i.e., $nnz_{\eta(j)} \geq nnz_{\eta(j+1)}$ for all $j < d^S$.
Consider indicator variables $\mathbb{I}(x) \in \{0,1\}^{d^S}$ for a datapoint $x^S$ which indicate whether the $j^{th}$ most popular dimension is nonzero, $\mathbb{I}(x)_j = x^S_{\eta(j)} \neq 0$.
Then the above algorithm is conceptually the same as sorting the indicator variables $\mathbb{I}(\bX)$ in decreasing order. In practice, we do not need to explicitly construct these indicator variables and sorting is performed by recursive partitioning. Thus the average complexity of cache sorting is essentially $O(N \log N)$. 

There are many conceivable modification to this basic approach (Algorithm.~\ref{algo:cache_sort}). For example, one can imagine performing sorting using gray-code sorting order~\cite{GrayCode}, by arranging the permuted indices of datapoints with $\mathbb{I}(x)_1=0~\text{AND}~\mathbb{I}(x)_2=1$ and $\mathbb{I}(x)_1=1~\text{AND}~\mathbb{I}(x)_2=0$ to be consecutive. In practice, modified variants often do not make a big differences in performance, and we used the simple decreasing order in our experiments. In addition, we use an optimized implementation of prefix sorting which, instead of sorting the original dataset, takes advantage the inverted index layout, requiring just 16 bytes per datapoint of temporary memory to efficiently compute the datapoint index permutation vector.

\subsection{Efficiency of Cache Sorting}
Let us view the {\em unsorted} sparse component $\bX^S$ as a sparse matrix with $N$ datapoints and $d^S$ dimensions, and suppose each entry of the matrix is independent. Let the $j^{th}$ dimension of $i^{th}$ datapoint, denoted as $x_{ij}$, is nonzero with probability $P_j$. Similarly the $j^{th}$ dimension of query vector $q^S$ is nonzero with probability $Q_j$.

Without loss of generality, let us assume that the dimensions are numbered as $1, 2, \cdots d^S$, and the probability that the entries of a datapoint $x\in \bX^S$ have nonzero values are sorted in decreasing order, i.e., $P_1 \geq P_2 \geq \cdots \geq P_{d^S}$.
For most real-world data, it is commonly observed that these probabilities tend to be distributed according to power law, i.e.,
\begin{equation*}
P_j~~\propto~~j^{~-\alpha} ~~~\textrm{for} ~~j=1, 2, \cdots d^S
\end{equation*}

If a block of $B$ values is completely zero, we can simply skip it.
Hence, the number of cache-lines we need to access for a given dimension $j$ can be computed as: $(1-(1-P_j)^B)\frac{N}{B}$. The expected number of cache-line accesses, $C_{unsort}$,
is:
\begin{equation}
\label{eqn:exp_unsort}
\E{[C_{unsort}]} = \sum_{j=1}^{d^S} Q_j (1-(1-P_j)^B) \frac{N}{B}
\end{equation}

After learning the permutation $\pi(\cdot)$ using the cache-sorting procedure from $\bX^S$, let us denote the matrix after permutation as $\bX'^{S}$. 
The number of contiguous blocks in dimension $j$ is $2^j$, and the average number of cache-lines each such block occupies is $\lceil \frac{P_j N}{2^j B} \rceil$.
In the worst case, no two
blocks can be fit in one cache-line. Then, an upper-bound on the expected number of cache-lines accessed is:
\begin{equation}
\label{eqn:exp_sorted}
\E{[C_{sort}]} \leq \sum_{j=1}^{d^S} Q_j \begin{cases}
    2^j \lceil \frac{P_j N}{2^j B} \rceil & \text{if $\frac{P_j N}{B} \geq 2^j$} \\
    (1-(1-P_j)^B) \frac{N}{B} & \text{otherwise}
  \end{cases}
\end{equation}

\begin{figure}
\begin{subfigure}{\columnwidth}
\centering

\includegraphics[width=0.8 \columnwidth]{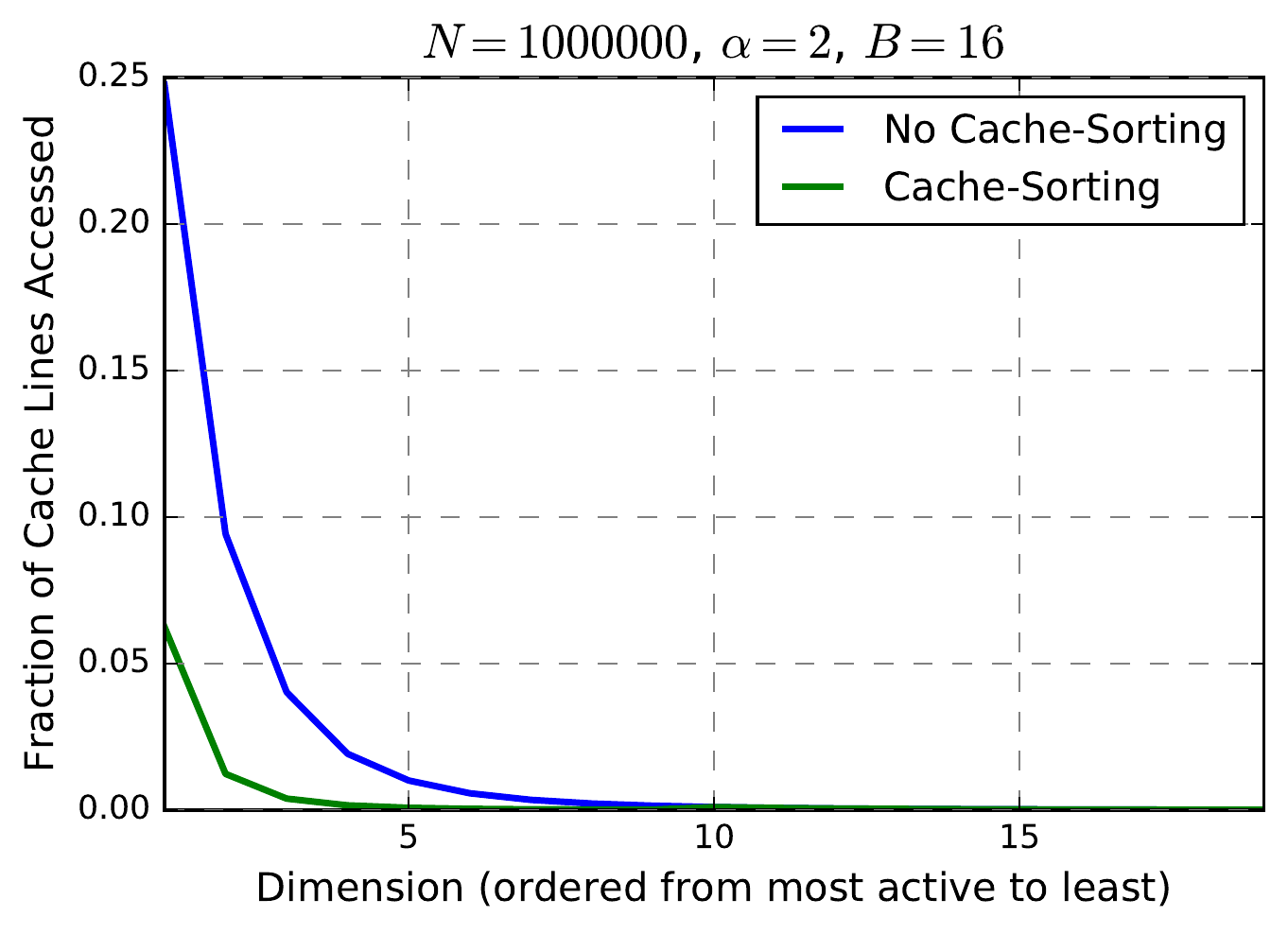}
\vspace{-.05in}
\caption{}
\label{fig:cachemiss1}
\vspace{-.02in}
\end{subfigure}

\begin{subfigure}{\columnwidth}
\centering
\includegraphics[width=0.8 \columnwidth]{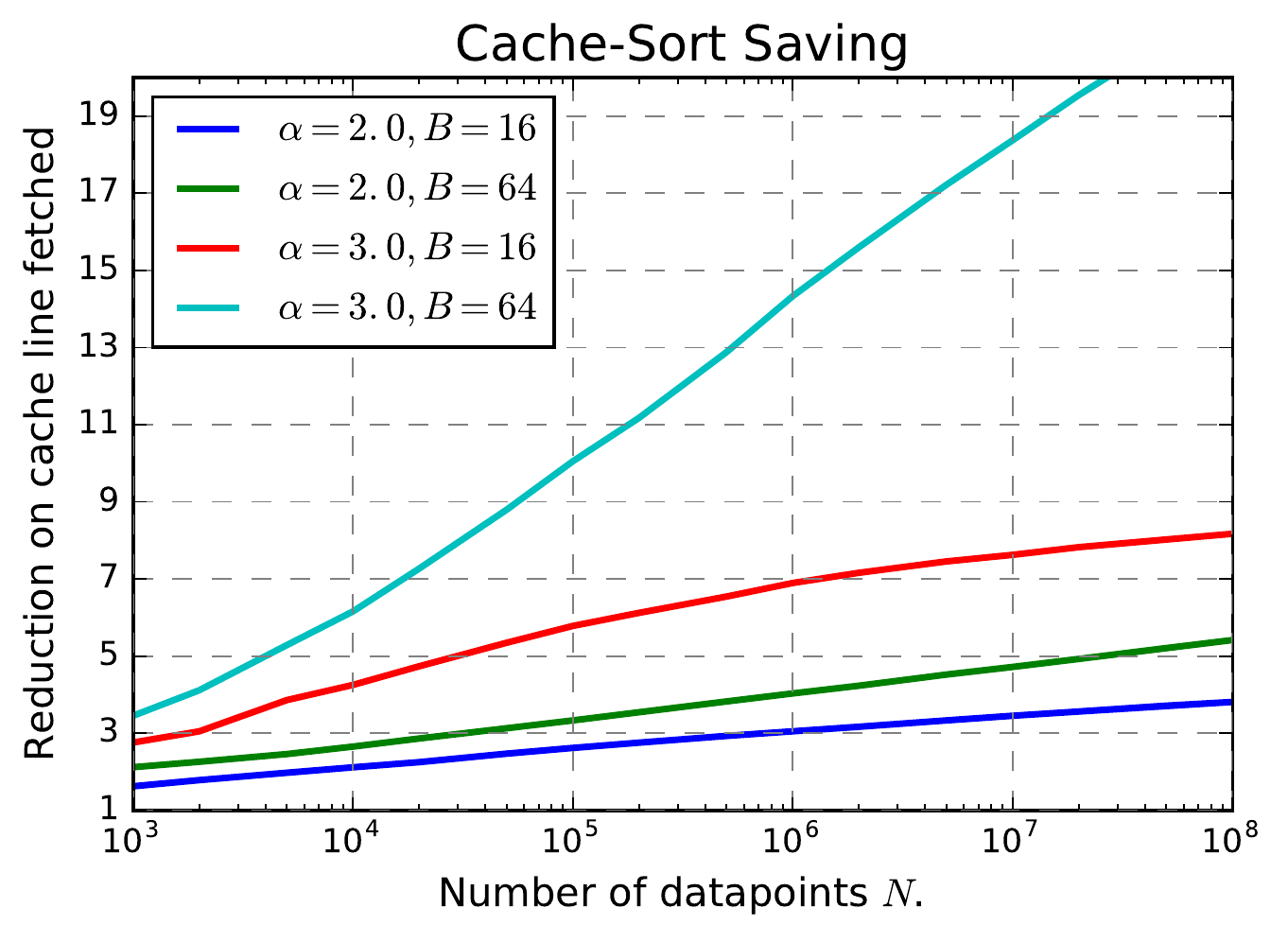}
\vspace{-.1in}
\caption{}
\vspace{-.07in}
\label{fig:cachemiss2}
\end{subfigure}
\caption{Savings in cache-line accesses using cache sorting. (a) Fraction of accumulator cache-lines accessed when computing inner products without and with cache-sorting, i.e., $\frac{\E{[C_{unsort}]}}{N / B}$, $\frac{\E{[C_{sort}]}}{N / B}$, respectively. (b) The number of times cache-sorting reduces cache-line access versus the unsorted, $\frac{\E{[C_{sort}]}}{\E{[C_{unsort}]}}$ as a function of $B, N$ and $\alpha$ Here, $B$ in $C_{unsort}$ is fixed to be 16.}
\vspace{-0.15in}
\label{fig:cachemiss}
\end{figure}

For simplicity, we assume $P_j=Q_j=j^{~-\alpha}$. 
In  Figure.~\ref{fig:cachemiss1}, we assume $N=1M$ datapoints, $\alpha=2.0$ and $B=16$ (64 byte cache-lines with floats) and plot the expected fraction of cache-lines accessed in the unsorted inverted index and the upper bound on the fraction in cache-sorted inverted index at each dimension $j$.
With cache-sorting, the expected number of cache-line accesses as well as the total expected number of accesses (area under the curve) drop significantly across all dimensions.

In Figure.~\ref{fig:cachemiss2}, we study the effect of parameters $N$, $\alpha$ and $B$. As $\alpha$ increases, the ``active'' dimensions will contain more datapoints, resulting in a larger cache-sorting impact.
The amount of saving also increases with cache-line size $B$, which motivates us to quantize values in the inverted index to reduce the memory footprint.

The analysis above is somewhat pessimistic as it assumes zero correlation between activity patterns of different dimensions. Most real-world datasets have significant correlations among dimensions e.g., several sets of dimensions tend to be active together. Empirically, we have observed over 10x improvement in throughput on several real-world datasets from proposed cache-sorting.
In addition to the benefits of reducing the total number of cache-lines read from main memory, for active dimensions, cache-sorting creates long sequential runs of nonzeros that can be encoded efficiently, and computed with vectorized SIMD instructions, further increasing the throughput.

\section{Indexing and Search}

\subsection{Dense Component}
\label{sec:indexing_dense}

As described in Section~\ref{sec:pq_intro}, we use product codes based approach for fast inner product approximation in dense space. 
Given Product Quantization (PQ) codebooks and a dense vector $x^D$, the indices of $K$ nearest centers that each of the $K$ subvectors $\{x^{D(k)}\}_{[K]}$ is assigned to are concatenated together into a sequence of integer codes.
If there are $l$ centers in each subspace codebook, the overall code length is $K\lceil \log_2 l \rceil$ bits.

\subsubsection{Asymmetric Distance Computation (ADC)} ADC allows us to compute the inner product between a quantized dense component $x^D$ and an {\em unquantized} query $q^D$ with high  efficiency and accuracy:

\begin{equation}
q^D \cdot x^D \approx q^D \cdot \phi_{PQ}(x^D) = \sum_{k \in [K]}
q^D(k) \cdot \phi_{VQ}(x^{D(k)}; \mathbf{U}^{(k)})
\nonumber
\end{equation}

Observe that in each subspace $k$, $\phi_{VQ}(x^{D(k)}; \mathbf{U}^{(k)})$ is essentially the index of one of the $l$ vectors from codebook $\mathbf{U}^{(k)}$. Thus, to approximate the subproduct $q^{D(k)} \cdot \phi_{VQ}(x^{D(k)}; \mathbf{U}^{(k)})$, one needs to simply precompute a {\em lookup table} $T(q^D, k)$ for each subspace based on the input query $q^D$. 
The subproduct in subspace $k$ is then a simple lookup in $T(q^D, k)[\textrm{index}(\phi_{VQ}(x^{D(k)}; \mathbf{U}^{(k)})]$.
The overall inner product is the sum of all $K$ subproducts.

\subsubsection{AVX2 In-Register Lookup Table}

An in-memory lookup table with $l=256$ codewords (LUT256) was proposed in \cite{PQ, OPQ, SQ}.
Our implementation uses $l=16$ codewords (LUT16), which is also reported in~\cite{SIMDLUT, AndreQuickADC, BlalockBOLT}.
While this results in slightly lower per-bit database encoding efficiency, LUT16 can be implemented via an in-register lookup table using the \texttt{PSHFB} x86 instruction.
This instruction was introduced in the SSSE3 instruction set, operating on 128-bit registers, and later extended to 256-bit registers in the AVX2 instruction set.
Each AVX2 \texttt{PSHUFB} instruction performs 32 parallel 16-way lookups of 8-bit values.
To perform accumulation without integer overflow, each register of 32$\times$8-bit values must be extended to $2$ registers of 16$\times$16-bit values, followed by $2$ \texttt{PADDW} instructions that perform parallel addition.

In this work, we present a significant optimization of the 8-bit to 16-bit width-extension operation, described as follows. First, we observe that we can implement width-extension of unsigned integers more efficiently than for signed integers. When constructing lookup tables, we bias the quantized lookup values from [-128,+127] to [0, 255], perform unsigned accumulation, then from the final values, we subtract the net effect of this bias.

Unsigned width extension (aka "zero-extension") from 1 register of 8-bit values into 2 registers of 16-bit values
can be accomplished using 2 instructions as follows.
Each 16-bit slot in the source register contains two of our 8-bit values ($R16 = 2^8 * R8_{high} + R8_{low}$).
The \texttt{PSRLW} instruction bit-shifts the source register ($R16_{high} = 2^8 * 0 + R16 / 2^8 = 2^8 * 0 + R8_{high}$), and the \texttt{PAND} instruction zeros out the upper bits ($R16_{low} = 2^8 * 0 + R8_{low}$).

We eliminate the \texttt{PAND} instruction, using the original register of 8-bit values as-is. This produces a result that is wrong by a large but known value; then we restore the correct value during a post-processing step.
This works despite the fact that the unzeroed bits cause our 16-bit accumulators to overflow many times. Overflows during addition are perfectly matched by a corresponding underflow during subtraction.

When operating on batches of 3 or more queries, our AVX2 implementation is able to sustain $\sim 16.5$ lookup-accumulate operations per clock-cycle on Intel Haswell CPUs, including the overhead for post-processing final inner products.
This is 8x better than a LUT256 implementation’s architectural upper-bound of two scalar loads per clock-cycle. 
When LUT16 operates on 1 query at a time, the CPU can compute inner products faster than it can stream the PQ codes from the main memory.
This constrains single-query performance to the maximum memory bandwidth available.

\subsubsection{Approximation Error from Product Quantization}

We first recap the rate distortion bound of parallel Gaussian channels \cite{CoverITBook} from information theory:
\begin{proposition}
The lower bound of the expected squared error (MSE) achievable with $b$ bits of quantization for $d$ i.i.d.~ Gaussian dimensions, each of which has variance $\sigma^2 / d$ is:
\[
\EE_x \|x^D - \tilde{x}^D\|_2^2 \geq \frac{\sigma^2}{2^{2b/d}}
\]

\label{lem:rate_distortion}
\end{proposition}

In practice, with the Lloyd's $k$-Means algorithm on Gaussian white noise~\cite{OPQ}, we can achieve an error within a few percentage of this theoretical lower bound.
For inner product computation, we can always whiten the dense component by multiplying $\bX^D$ with the matrix $P=\textrm{Cov}^{-1/2}(\bX^D)$.
At query time, $q^D$ is also multiplied by $(P^{-1})^T$.
This is a special case of QUIPS~\cite{QUIPS}, where the query distribution is the same as datapoint distribution.
So from this point on, we assume that there exists a small constant $c$, such that with suitable seeding, we can always achieve an MSE $\le (1+c) \sigma^2 2^{-2b/d^D}$.
We then present following bound using Azuma's inequality :

\begin{proposition}
The difference between the exact inner product $q^D \cdot x^D$ and its approximation $q^D \cdot \tilde{x}^D$ is bounded by $\epsilon$ with probability:
\[
\textrm{Pr}_q\{ |q^D \cdot x^D - q^D \cdot \tilde{x}^D| < \epsilon \} \geq 1 -
2e^{-\frac{\epsilon^2}{2 K \max_k \|q^{D(k)}\|^2_2 \max_{x, k} \|x^{D(k)} - \tilde{x}^{D(k)}\|^2_2}}
\]
Where $q^{D(k)}$, $x^{D(k)}$ and $\tilde{x}^{D(k)}$ are the $k^{th}$ subvectors of $q^D$, $x^{D}$ and $\tilde{x}^{D}$, respectively under product quantization. The probability is defined over the randomness in the query.

\label{thm:dense_probabilistic_bound}
\end{proposition}

We also note that the upper bound $\max_{x, k} \|x^{D(k)} - \tilde{x}^{D(k)}\|^2_2$ is taken over all datapoint $x$ and all subspace $k$.

\subsection{Sparse Component}

We can generate an even sparser representation of an already sparse dataset by {\em pruning} out small entries while maintaining accurate approximation to inner products.
Pruning reduces the size of the index and the amount of arithmetic operations during search.
An even sparser data also helps cache sorting generate even more continuous memory layout of the index.

We first observe that on average only a small number of dimensions contribute nonzero products to the overall inner product between two sparse vectors.
Also, in each dimension, the occurrence of large absolute values is rare across a dataset.
These two phenomena combined make the contribution from small entries to the overall inner product much less significant than that from large entries, especially if our goal is to identify largest inner products.
Mathematically, we represent a sparse vector as $x^S=\{(j, x^S_j) : x^S_j \neq 0, j \in d^S\}$.
And given a set of thresholds $\{\eta_j\}$, one for each dimension, pruning outputs a sparser representation as: $\textrm{Prune}(x^S; \{\eta_j\}) = \{(j, x^S_j) : |x^S_j| \geq \eta_j, j \in [d^S]\}$.

\subsubsection{Approximation Error from Pruning}

We now investigate the error in inner product approximation due to pruning.
We proceed with the assumption that {\em both} $q^S$ and $x^S$ are generated by the following process:

\begin{itemize}
\item Each dimension is independently nonzero with probability $p$ and zero with probability $1-p$.
\item If nonzero, the value for dimension $k$ is sampled from a distribution $F_k$,
and the support of $F_k$ is $[-M, M]$.
\end{itemize}

The error of pruning can be characterized with Chernoff bound:

\begin{proposition}
The difference between the exact inner product $q^S \cdot x^S$ and its approximation $q^S \cdot \tilde{x}^S$ is bounded by $\epsilon$ with  probability:

\[
\textrm{Pr}_x\{|q^S \cdot x^S - q^S \cdot \tilde{x}^S| < \epsilon\} \geq
1 - 2e^{-\frac{(n_\epsilon - dp^2)^2}{n_\epsilon + dp^2}}
\]

Where $\tilde{x}^S$ is the pruning output of $x^S$ and $n_\epsilon = \epsilon / (M\max \eta_j)$.

\label{thm:sparse_probabilistic_bound}
\end{proposition}

The probability of two vectors having nonzero values for the same dimension is typically low, i.e., $dp^2 \ll 1$.
So the above probability approaches 1 asymptotically at the rate of $O(e^{-n_\epsilon})$, where $n_\epsilon$ is the minimum number of entries that need to be removed from the sparse component to cause an error bigger than $\epsilon$.

\section{Residual Reordering for Improved Search}
\label{sec:residual}

The index that we build provides a lossy approximation of the original component, and we define {\em residual} as the difference between the original component and its index approximation.
When retrieving $h$ datapoints from the dataset, we can overfetch $\alpha h$ candidates from the original data index and then use the residual to reorder these overfetched candidates.
Due to the simple decomposition $q \cdot x = q \cdot \tilde{x} + q \cdot (x - \tilde{x})$, we can obtain the exact inner product by adding the inner product between the query and the residual to the approximate inner product from the index.
With the exact inner product, we can refine the $\alpha h$ candidates and improve our accuracy for top $h$ candidates significantly.
This allows us to reap the benefit of fast index scanning described in previous sections and also maintain high recall at a small cost due to residual reordering.
To further reduce the cost of this residual reordering step, we can also build another index on the residual, which is much more precise than the data index.
We apply this paradigm multiple times to achieve the best trade-off between speed and accuracy:

\begin{enumerate}

\item \textbf{Overfetch $\alpha h$ from both sparse and dense data indices}:
we take the sum of both approximate inner products from sparse and dense components and retain the largest $\alpha h$ for further reordering.
The parameter $\alpha > 1$ can be tuned to balance recall and search time.

\item \textbf{Reorder with dense residual index and retain $\beta h$}:
we perform residual reordering with the dense residual index for the $\alpha h$ remaining datapoints and only retain $\beta h$ for the last reordering.
The parameter $\beta > 1$ also needs to be tuned.

\item \textbf{Reorder with sparse residual index and return $h$}:
we perform residual reordering with the sparse residual index for the $\beta h$ remaining datapoints and return the largest $h$ as the final search result.
\end{enumerate}

Reordering is always performed on a small subset of $O(h)$ residuals.
And its run time is much shorter in practice when compared with the run time of scanning through the sparse and dense data indices.
In our similarity search benchmarks, the residual reordering logic consumes less than $10\%$ of the overall search time.
On the other hand, it improves the final retrieval recall significantly.

\subsection{Retrieval Performance}
\label{sec:retrieval_performance}

In the context of similarity search, rank preservation is of great interest and our hybrid approach with residual reordering gives following retrieval performance guarantee:

\begin{proposition}

Given a fixed query $q$, define the $(h, \alpha)$ gap as $G_{h, \alpha h} = q \cdot x^*_k - q \cdot x^*_{\alpha h}$, the proposed hybrid search approach achieves a \texttt{recall@h} of at least $r$ as long as following condition holds:

\[
\textrm{Pr}_x \{ |q \cdot x - q \cdot \tilde{x}| < G_{h, \alpha h} / 2 \} \geq r
\]

Where $x^*_h$ is the $h^{th}$ datapoint in terms of inner product and $x^*_{\alpha h}$ is the $\alpha h^{th}$ datapoint.
The probability is defined over the randomness in datapoint generation.

\label{thm:retrieval_performance}
\end{proposition}

This proposition states the fact that recall is at least the probability of the event that inner product approximation error is less than half the gap between $h^{th}$ and the $(\alpha h)^{th}$ largest inner products.
Based on the bounds from Proposition~\ref{thm:dense_probabilistic_bound} and \ref{thm:sparse_probabilistic_bound}, we can almost always find a large enough $\alpha$ that satisfies this condition so long as the inner product distribution exhibits a quickly vanishing upper tail.
For most real world datasets, under the setting of $h \ll N$, $\alpha$ is empirically $\le 10$ to achieve $\geq 90\%$ recall.

\section{Overall Indexing Algorithm}
To efficiently approximate inner products between datapoints in the dataset $\bX$ and a query $q$, we index the dataset $\bX$ as follows:

\begin{enumerate}
\item Build an index for the sparse component by pruning with per dimension threshold $\eta_j$:
\begin{equation}
\textrm{Prune}(x^S; \{\eta_j\}) = \{(j, x^S_j) : |x^S_j| \ge \eta_j, \, j \in [d^S] \}
\label{eq:sparse_sketch_1}
\end{equation}
The residual $R^S(x^S)$ is thus $x^S - \textrm{Prune}(x^S; \{\eta_j\})$.
To efficiently compute the inner product between the query and the residuals, we also build another pruned index of the residuals with parameter $\epsilon_j$.
\begin{equation}
\textrm{Prune}(R^S(x^S); \{\epsilon_j\}) = \{(j, x^S_j) : \eta_j > |x^S_j| \ge \epsilon_j, \, j \in [d^S] \}
\label{eq:sparse_sketch_2}
\end{equation}

\item Build an index for the dense component by applying product quantization (PQ) with codebooks $\{\mathbf{U}^{(k)}\}$ for $K_U$ subspaces.
\begin{equation}
\textrm{PQ}(x^D; \{ \mathbf{U}^{(k)} \}) = \phi_{PQ}(x^D; \{\mathbf{U}^{(k)}\}_{[K_U]})
\nonumber
\end{equation}
Similar to sparse indexing, we build another PQ index of the residuals $R^D(x^D) = x^D - \textrm{PQ}(x^D; \{ \mathbf{U}^{(k)} \})$ with codebooks $\{\mathbf{V}^{(k)}\}$ for $K_V$ subspaces as:
\begin{equation}
\textrm{PQ}(R^D(x^D); \{ \mathbf{V}^{(k)} \}) = \phi_{PQ}(R^D(x^D); \{\mathbf{V}^{(k)}\}_{[K_V]} )
\nonumber
\end{equation}
\end{enumerate}

\subsection{Parameter Selection}

\subsubsection{Dense Index Parameters}

The first dense data index is built with $K_U$ codebooks $\{\mathbf{U}^{(k)}\}$ designed to have a relatively low bit rate, i.e., $K_U = d^D / 2$ and $l=16$, which means we store 4 bits for on average 2 dense dimensions.
This immediately achieves a data index size reduction of 16x when compared to the original data that requires 32 bits per dimension for a floating point number.
And it also enables us to apply in-register table lookup (LUT16) that outperforms alternative arithmetic or in-memory table lookup operations by at least 4x.
The second residual index is built with $K_V = d^D$ and $l=256$.
Since now we treat each dimension as a subspace, we can directly apply scalar quantization with a distortion of at most $1/256$ of the dynamic range.
This residual index is exactly $1/4$ the size of the original dataset and its error in inner product approximation is unnoticeable for our tasks.

\subsubsection{Sparse Index Parameters}

For fast approximation, we normally set the first threshold $\eta_j$ from Equation~\ref{eq:sparse_sketch_1} to a relatively high value so that only top $100$s of nonzero values in dimension $j$ are kept in the data index.
This high threshold reduces the size of the data index to about 2 orders of magnitude smaller than the original dataset, and hence achieves significant speedup in index scan.
Empirically, this hyper-sparse index induces little loss in retrieval performance.
The second threshold $\epsilon_j$ from Equation~\ref{eq:sparse_sketch_2} is set to a relatively low value so that most of the nonzero values are kept in the residual index.
The reason is that we only need to access the residual index for a moderate number of datapoints, so its size does not affect computation speed directly like the size of the data index.
With this relatively accurate residual index, the sum of the data index approximation and the residual index approximation is almost exact, which helps lift recall to high 90s\%.

\section{Experiments}
\label{sec:experiments}

\subsection{Evaluation Datasets}
\label{sec:datasets}
In this work, we use two public datasets and one large scale industrial dataset for experiments. 

\subsubsection{Public Datasets}

We use hybrid versions of two public datasets Netflix~\cite{BennettNetflixPrize} and Movielens~\cite{HarperMovielens}.
Each dataset contains a sparse collection of user-movie-rating triplets.
Our goal is to look for users in the dataset that have similar movie preferences as the users in the query set.
The sparse component of a user comes directly from the set of ratings that the user provides.
For dense components, we first assemble the user-movie-rating matrix $M$, whose rows represent users and columns represent movies, and values are the ratings from 1 to 5.
We follow the classic collaborative filtering approach~\cite{BillsusSVD} and then perform Singlar Value Decomposition on the sparse matrix $M \approx USV^T$.
The dense components of all users are given by $U$ weighted by $\lambda$ .
So the combined hybrid vector representation of users is $\bigl( \lambda U|M \bigr )$, in which each row is a user embedding.
The dense component dimensionality is fixed to be 300.  Other details can be found in Table~\ref{tbl:perfcomp}.
We randomly sample 10k embeddings as the query set and the rest as the dataset.

\subsubsection{QuerySim Dataset}
We introduce a hybrid dataset that is built from a large set of most frequent queries in web search for the task of identifying similar queries.
Each query is processed through two pipelines to generate a sparse and a dense component of its vector representation.
The sparse pipeline generates a sparse vector, representing the bag of unigrams and bigrams in the query and each gram's weight is set to a customized $\textrm{tf}\times\textrm{idf}$ value.
The dense pipeline generates a dense vector representation using a neural network similar to~\cite{LeDoc2Vec}. 
We fine tuned the relative weights for the sparse and dense inner products to optimize a more comprehensive metric such as the area under ROC.	 
Finally, all datapoints are scaled with the learned relative weights and stored in a database.

\begin{figure}
\begin{subfigure}[h]{0.48\linewidth}
\includegraphics[width=\linewidth]{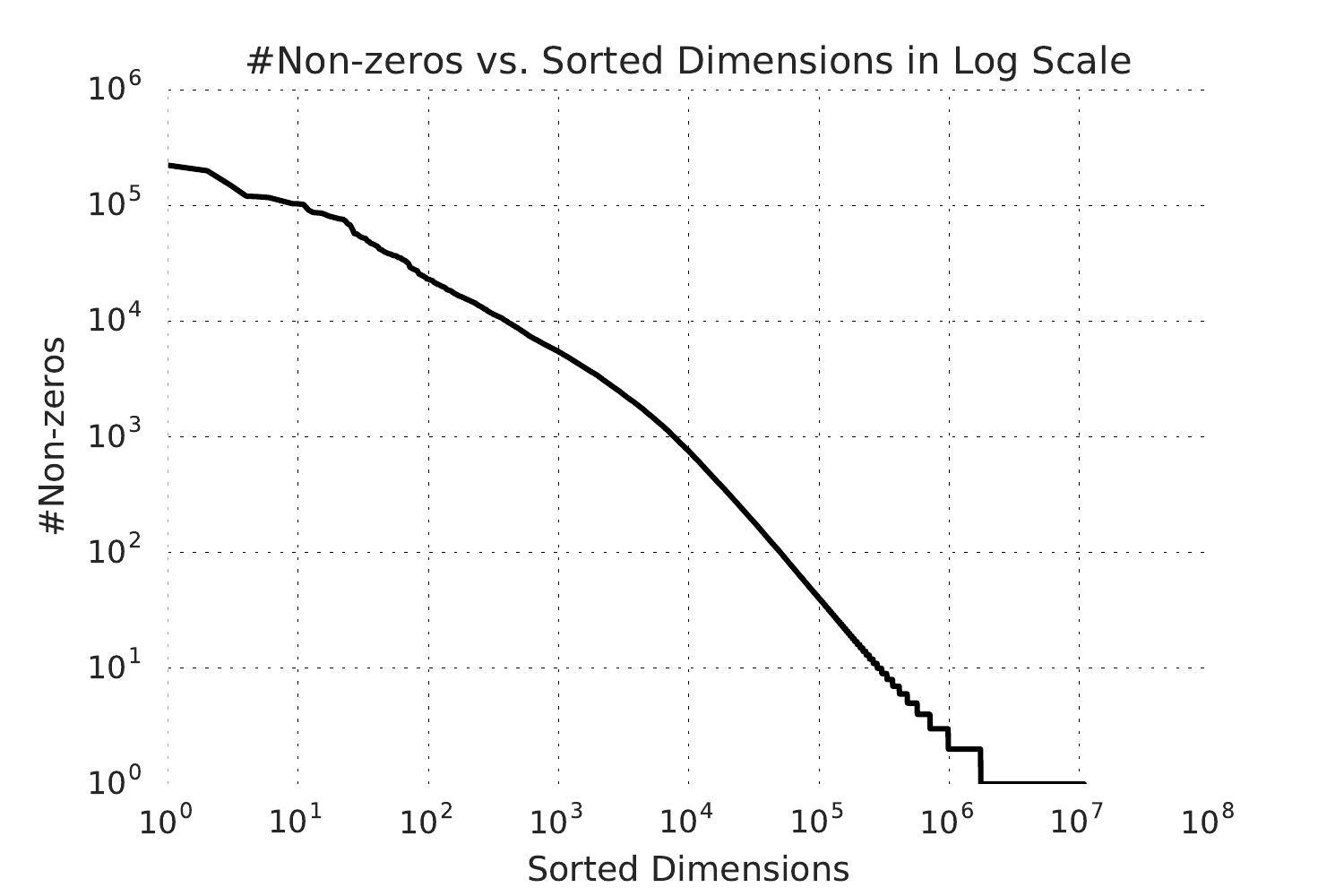}
\caption{}
\label{fig:distribution_cols}
\end{subfigure}
\hfill
\begin{subfigure}[h]{0.48\linewidth}
\includegraphics[width=\linewidth]{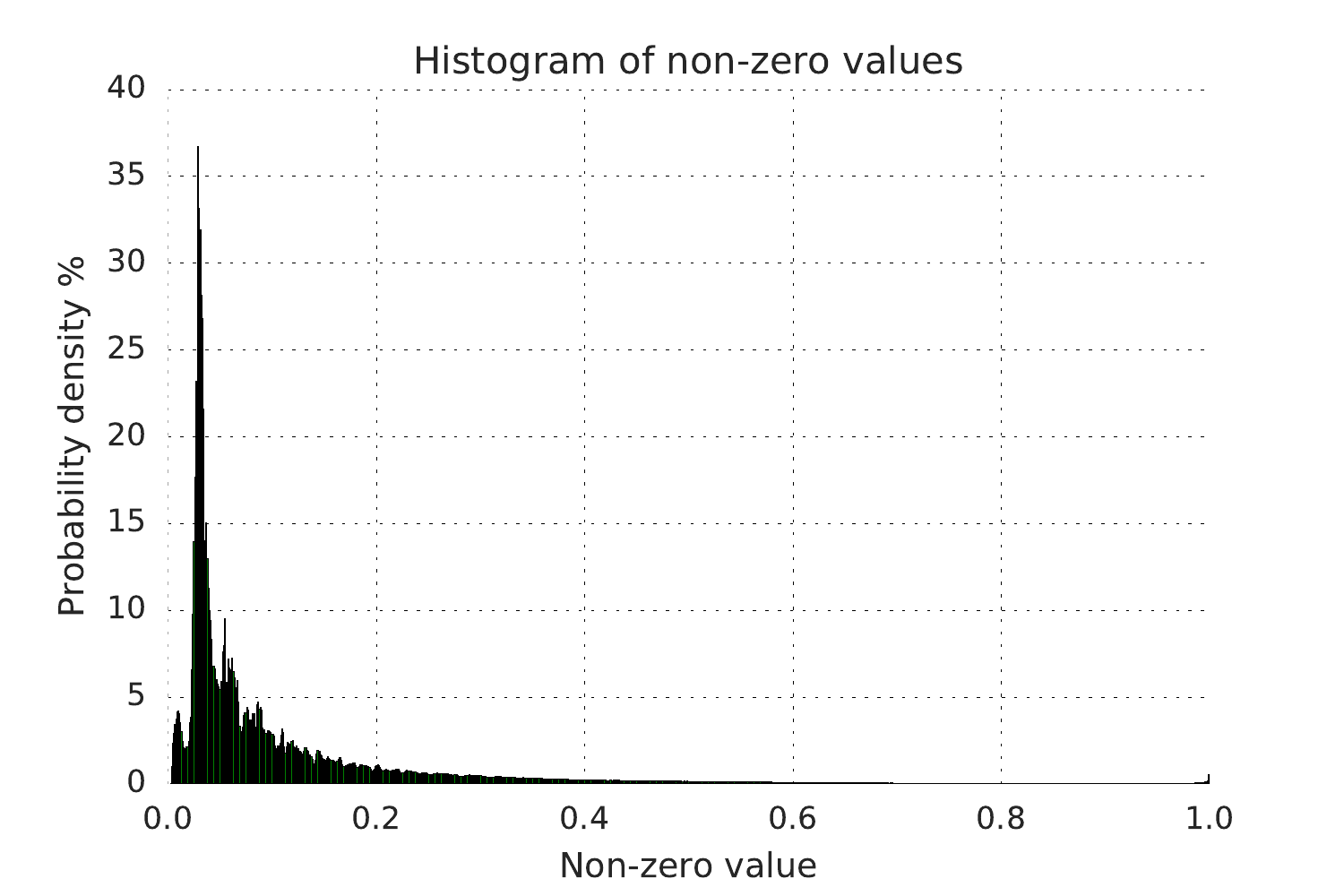}
\caption{}
\label{fig:sparse_value_hist}
\end{subfigure}%
\caption{Statistics on the sparse component of the QuerySim dataset: (a) the power-law distribution of the number of nonzeros for sorted dimensions in log-scale, and (b) the histogram of the nonzero values with median 0.054, 75\% percentile 0.12, and 99\% percentile 0.69,  which justifies the pruning strategy for sparse index.}
\vspace{-0.15in}
\end{figure}

The overall dataset contained one billion points and the dimensionality of the hybrid vector was over 1 billion (Table~\ref{tbl:hybrid_datasets}).
Zooming into the sparse component, we first demonstrate the power-law distribution of the numbers of nonzeros across dimensions in Figure~\ref{fig:distribution_cols}.
We then visualize the long tail in the distribution of values of nonzero entries in the sparse component in Figure~\ref{fig:sparse_value_hist}.

\begin{table}
\begin{tabular}{| l | l | l |}
\hline
\#datapoints & \#Dense Dims & \#Active Sparse Dims \\
\hline
$10^9$ & 203 & $10^9$ \\
\hline
\#Avg Sparse nonzeros & On-disk Size & Update Frequency \\
\hline
 134 & 5.8TB & Weekly \\
\hline
\end{tabular}
\caption{QuerySim dataset used for the sparse-dense hybrid similarity search benchmark.}
\vspace{-0.25in}
\label{tbl:hybrid_datasets}
\end{table}

\subsection{Recall and Speedup}
\label{sec:speedup}

Our goal is to compare the speed and recall of top 20 items from the entire dataset for each technique. 
All single machine benchmarks are conducted on a workstation with an Intel CPU clocked at 3.5GHz, with 64GB main memory.
Distributed benchmarks are performed on a cluster with 200 servers that are equipped with an Intel CPU clocked at 2GHz and 128GB main memory.
Both CPUs support the AVX2 instruction set.

For comparison we use the following well-known baselines: 

\textbf{Exact methods}:
with {\em Dense Brute Force}, we pad $0$'s to the sparse component to make the dataset completely dense;
with {\em Sparse Brute Force}, we append the sparse representation of the dense component to the end of the sparse component to make the dataset completely sparse.
With {\em Sparse Inverted Index}, we perform the same conversion as sparse brute force and search with an inverted index instead.

\textbf{Hashing methods}:
with {\em Hamming (512 Bits)}, we first project each datapoint onto 512 Radmacher vectors and binarize the projected values with median thresholding.
We use these 512-bit binary codes to perform hamming distance search and retrieve top 5K points, from which the required 20 are retrieved via exact search.

\textbf{Dense only methods}:
with {\em Dense PQ, Reordering 10k}, we
apply PQ indexing to only the dense component.
From the index, we fetch top 10k datapoints and then return top 20 via exact search.

\textbf{Sparse only methods}:
with {\em Sparse Inverted Index, No Reordering}, we simply retrieve top 20 from the sparse component with an inverted index and return them.
with {\em Sparse Inverted Index, Reordering 20k}, we retrieve top 20k from the sparse component with an inverted index and then compute exact inner products of the 20k and return top 20.

Table~\ref{tbl:perfcomp} shows the comparison of different methods on public datasets. It is clear that the proposed method gives very fast search with high recall.
The results on our large scale QuerySim dataset are summarized in Table~\ref{tbl:sparse_dense_only}.
The latency and recall are measured on a random sample of 5M datapoints from the original 1B datapoints in order to compare with other more expensive techniques.  
The proposed hybrid approach is over 20x faster than the closest competition in the high recall regime.

\textbf{Scalability}: Extrapolating these results to all-pair search scenario (i.e., both query and database sets are the same with 1B billion datapoints), with $10^4$ CPU cores, the exact sparse brute force methods will take about \textbf{9 years} to complete the 1B$\times$1B search on the QuerySim dataset while sparse inverted index will take about \textbf{3 months}. On the other hand, our proposed hybrid approach is able to complete this under \textbf{1 week} to match the weekly update requirement from production.

\textbf{Online Search}: To support online retrieval applications with stringent latency requirement, we have also developed a distributed similarity search system.
We divide the dataset randomly into 200 shards, and allocate 200 servers, each of which loads a single shard into memory.
For a single query, our system achieves $90\%$ \texttt{recall@20} at an average latency of \textbf{79ms}.

\begin{table}
\centering
\begin{tabular}{|p{3.0cm}|l|l|l|l|}
\hline
Dataset $\rightarrow$       & \multicolumn{2}{l|}{Netflix Hybrid} & \multicolumn{2}{l|}{Movielens Hybrid} \\ \hline
\#datapoints         & \multicolumn{2}{l|}{$5 \times 10^5$}            & \multicolumn{2}{l|}{$1.4 \times 10^5$}                 \\ \hline
\#Dims                & Dense            & Sparse           & Dense             & Sparse            \\ \hline
                      & 300              & $1.8 \times 10^4$            &  300                 &  $2.7 \times 10^4$                 \\  \hline
\textbf{Algorithm $\downarrow$}             & \textbf{Time}             & \textbf{Recall}           & \textbf{Time}              & \textbf{Recall}            \\ \hline
\textbf{Dense Brute Force}            & 3464             & 100\%            & 1242              & 100\%             \\ \hline
\textbf{Sparse Brute Force}             & 905              & 100\%            & 205               & 100\%             \\ \hline
\textbf{Sparse Inverted Index} &  63.9             & 100\%            & 15.7              & 100\%             \\ \hline
\textbf{Hamming (512 bits)}    & 16.0               & 9\%              & 11.5              & 20\%              \\ \hline
\textbf{Dense  PQ, Reordering 10k} & 52.2 & 98\% & 29.4 & 100\% \\ \hline
\textbf{Sparse Inverted Index, No Reordering} & 22.8 & 29\% & 5.1 & 98\% \\ \hline
\textbf{Sparse Inverted Index, Reordering 20k} & 96.8 & 70\% & 49.0 & 100\% \\ \hline
\textbf{Hybrid (ours)}                 & \textbf{18.8}    & \textbf{91\%}    & \textbf{2.6}     & \textbf{92\%}     \\ \hline
\end{tabular}
\caption{Hybrid search performance with public datasets.
All timings are reported in ms (per query) and recall measured at top 20.}
\label{tbl:perfcomp}
\vspace{-0.15in}
\end{table}

\begin{table}
\centering
\small
\begin{tabular}{|p{5cm}|l|l|}
\hline
Dataset      & \multicolumn{2}{l|}{QuerySim} \\ \hline
\#datapoints         & \multicolumn{2}{l|}{$5 \times 10^6$ (sampled)} \\ \hline
\#Dims                & Dense            & Sparse     \\ \hline
                      & 203              & $10^9$      \\ \hline
\textbf{Algorithm}           & \textbf{Time (ms)}             & \textbf{Recall@20} \\ \hline  
\textbf{Dense Brute Force}                                     & OOM                                        & OOM                          \\ \hline
\textbf{Sparse Brute Force}                                     & 9655                                        & 100\%                          \\ \hline
\textbf{Sparse Inverted Index}                                  & 406                                         & 100\%                          \\ \hline
\textbf{Hamming (512 bits)}   & 59.5                                        & 0\%                          \\ \hline
\textbf{Dense PQ, Reordering 10k}                  & 39.8                                         & 45\%                         \\ \hline
\textbf{Sparse Inverted Index, No Reordering}   & 58.6                      & 0\%           \\ \hline
\textbf{Sparse Inverted Index, Reordering 20k}     & 102                    & 30\%          \\ \hline
\textbf{Hybrid (ours)}  & \textbf{20.0}            & \textbf{91\%} \\ \hline
\end{tabular}

\caption{Hybrid search performance with a sampled 5M datapoints from the QuerySim dataset .
The dense brute force method runs out of memory due to billion dimensionality.}
\label{tbl:sparse_dense_only}
\vspace{-0.3in}
\end{table}

\section{Conclusion}
\vspace{2mm}
We have introduced a novel yet challenging search scenario for performing efficient search in high dimensional sparse-dense hybrid spaces.
To achieve high performance and accuracy, we proposed a  fast technique based on approximating inner product similarity. Extensive optimizations to in-memory index structures for both sparse and dense vectors were also described to take advantage of modern computer architecture. We have shown advantages of our approach in large-scale real-world search scenario, which achieves more than 10x speedup with recall in 90s\%. 

With this fast and accurate approximation method, we hope that novel information retrieval systems that operate directly on sparse and dense features can be developed to unlock the value of hybrid data at massive scale. 

\tiny
\bibliographystyle{ACM-Reference-Format}
\bibliography{kdd18}

\end{document}